%% file: main.tex
\documentclass[letterpaper, 10pt, conference]{ieeeconf}
\pdftrailerid{}
\IEEEoverridecommandlockouts
\usepackage{graphicx}
\usepackage{amsmath}
\usepackage{amssymb}
\usepackage{booktabs}
\usepackage{multirow}
\usepackage{xcolor}
\usepackage{xspace}
\usepackage{url}
\usepackage{algorithm}
\usepackage[noend]{algpseudocode}
\usepackage{pgfplots}
\pgfplotsset{compat=1.18}
\usepgfplotslibrary{groupplots}
\usetikzlibrary{arrows.meta,positioning,fit,backgrounds}
\newcommand{\ours}{\texttt{vla.simd}\xspace}
\newcommand{\feff}{f_{\mathrm{eff}}}
\newcommand{\fc}{f_{\mathrm{c}}}
\newcommand{\Tq}{T_{\mathrm{q}}}
\newcommand{\Pf}{P_{\mathrm{fma}}}
\newcommand{\Pl}{P_{\mathrm{ld}}}
\newcommand{\Rmax}{R_{\max}}
\newcommand{\mr}{m_{\mathrm{r}}}
\newcommand{\nr}{n_{\mathrm{r}}}
\newcommand{\nv}{n_{\mathrm{v}}}
\newcommand{\fail}[1]{\textcolor{black!45}{#1}}
\newcommand{\cone}[1]{{\normalfont\itshape #1}}
\title{\LARGE \bf
\texttt{vla.simd}: Efficient CPU Inference\\
for Language-Conditioned Manipulation
}

\author{Khanh D. Nguyen$^{1}$, Hoang M. Truong$^{1}$, An T. Le$^{1,2,3}$%
\thanks{$^{1}$VinRobotics, Vietnam.}%
\thanks{$^{2}$Center for AI Research, VinUniversity, Vietnam.}%
\thanks{$^{3}$Intelligent Autonomous Systems, TU Darmstadt, Germany.}%
\thanks{Corresponding author: An T. Le.}%
}

\begin{document}

\bstctlcite{IEEEexample:BSTcontrol}
\maketitle
\thispagestyle{empty}
\pagestyle{empty}

% =====================================================================
\begin{abstract}
% =====================================================================
Deploying language-conditioned manipulation without a dedicated GPU requires efficient inference and action chunks that cover the delay between policy queries.
We present \ours{}, a CPU inference engine that combines shared SIMD micro-kernels, reusable computation, and target-specific optimization.
We relate query latency and execution horizon to action availability under lagged and time-aligned execution, distinguishing action supply from feedback frequency.
Across six policies and four CPUs, \ours{} achieves approximately $1.4\times$ median speedup over compiled PyTorch references while preserving fp32 numerical fidelity.
We also introduce IMPACT, an ACT-based policy with cached text representations and language-modulated visual features.
IMPACT is the only language-conditioned policy in our evaluated set that supplies at least 30 actions/s on the Raspberry Pi~5: after a 90\,s thermal soak, it supplies 33.5 actions/s in fp32 and 81.2 with int8.
Separate GPU evaluations yield 76.4\% mean success across four LIBERO suites without robot pretraining; instruction-shuffling tests demonstrate selection among familiar goals.
Trials with IMPACT on an SO-101 arm and SmolVLA on a UR10e with a Robotiq gripper demonstrate CPU deployment on two robot embodiments.
Project page: \url{https://vla-simd.github.io/}.
\end{abstract}

% =====================================================================
\section{Introduction}
% =====================================================================

Deploying a language-conditioned manipulation policy~\cite{openvla2024,pi0,octo2024,smolvla2025} on a robot without a dedicated GPU raises a practical question: can the available CPU deliver actions fast enough to keep the robot moving?
Answering this requires more than comparing model sizes or query latencies.
A manipulation policy must provide actions within the timing constraints of its controller while retaining the ability to select the goal specified by the user.

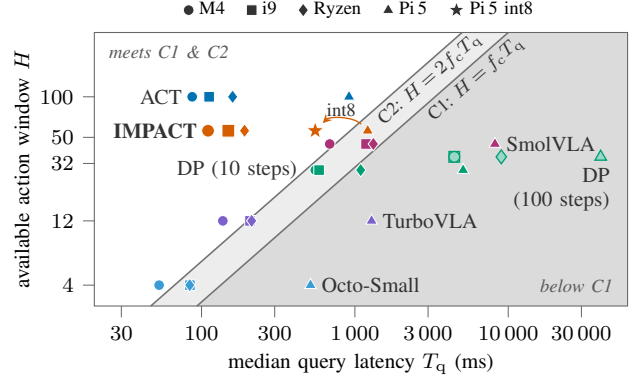
\begin{figure}[t]
\centering
\input{figs/budget.tex}
\caption{Nominal action-supply budgets of chunked policies.
Markers show each policy's available action window $H$ and median query latency $\Tq$ on each CPU.
Rows sharing $H$ are offset by a factor of 1.12; no marker crosses a line.
The lines omit control-step rounding: above C1 a policy supplies actions at least as fast as a $\fc=30$\,Hz controller consumes them; above C2 it still does when stale actions are discarded (Section~\ref{sec:budget}).
On the Pi~5, ACT meets C1 with 100 actions per query, whereas TurboVLA fails with 12; IMPACT meets C1 in fp32 and C2 with int8.}
\label{fig:budget}
\end{figure}

Action chunking creates an opportunity for CPU deployment.
By predicting several future actions in each query~\cite{act2023,diffusionpolicy2023}, a policy can overlap inference with execution and operate below the controller's update frequency~\cite{rtc2025,smolvla2025}.
The relevant constraint is therefore whether the next usable action chunk arrives before the current one is exhausted.
This depends jointly on inference latency, execution horizon, and how delayed actions are handled (Section~\ref{sec:budget}, Fig.~\ref{fig:budget}).
A faster query does not necessarily provide a higher action-supply rate, and meeting a nominal compute budget does not by itself guarantee responsive or successful manipulation.
CPU deployment must therefore meet the action-supply budget while preserving language-based goal selection.

\ours{} combines CPU optimization with action-supply analysis to identify which policies can sustain execution on each processor.
Linear layers, convolutions, and attention projections share matrix computations that can exploit SIMD instructions through packed-panel micro-kernels~\cite{goto2008anatomy,vanzee2015blis}.
Data layout, register use, memory reuse, and threading must suit the policy and processor: a setting that helps one CPU can hurt another.
Our engine, \ours{}, combines shared SIMD micro-kernels with target-specific choices and separates \emph{computation lifetimes}: weights are packed per process, instruction-only tensors are cached per episode, and observation-dependent features are computed per query.

Efficient execution must also preserve the information needed to choose the task.
ACT~\cite{act2023} provides an efficient starting point, meeting the basic 30\,Hz action-supply budget on the Pi~5, but it does not condition on language instructions.
In a multi-task setting, the same scene can correspond to different goals, such as placing the tape into either the box or the cup; the observation alone cannot specify the user's intention.
We therefore introduce IMPACT (Instruction-Modulated Perception and Action Chunking with Transformers), an action-chunking policy that extends ACT with language conditioning through cached text representations, instruction tokens, and visual modulation.

Our contributions are:
\begin{itemize}
\item \textbf{\ours{}}, a CPU engine with a shared SIMD micro-kernel for linear layers, convolutions, and attention projections, plus target-specific tiling and quantization.
\item \textbf{IMPACT}, a language-conditioned action-chunking policy whose language conditioning accounts for approximately 2.6\% of per-query compute in the SO-101 configuration, evaluated through LIBERO, instruction shuffling and real SO-101 manipulation.
\item \textbf{A study of six policies on four CPUs and two embodiments}, showing where CPU optimization meets action-supply budgets, while assessing thermal effects, PyTorch fidelity, and robot deployment.
\end{itemize}

% =====================================================================
\section{Related Work}
% =====================================================================

\textbf{Efficient vision-language-action policies.}
Billion-parameter vision-language-action (VLA) models~\cite{openvla2024,pi0,pi05} motivated compact designs that shorten or remove the language model~\cite{smolvla2025,turbovla2026,efficientvla2025}, reduce the overall parameter count~\cite{tinyvla2025,nanovla2025}, or quantize the policy~\cite{bitvla2025}; see~\cite{efficientvlasurvey2025} for a survey.
Recent studies characterize VLA latency on GPUs and embedded accelerators~\cite{vlaperf2026,edgechar2026}, with broader hardware comparisons also including CPUs~\cite{xpuchar2026}.
Lite VLA reports CPU deployment on a Pi~4, with inference taking seconds to minutes~\cite{litevla2025}.
We compare six policies on four CPUs using the same timing boundary and account for the actions supplied by each query.

\textbf{Chunked execution in real time.}
ACT~\cite{act2023} and Diffusion Policy~\cite{diffusionpolicy2023} predict action chunks.
Real-time chunking~\cite{rtc2025} overlaps inference with motion and guides new chunks for continuity; LeRobot supports asynchronous serving~\cite{smolvla2025,lerobot2026}.
Bidirectional decoding~\cite{bid2025} addresses the consistency--reactivity trade-off through guided sampling.
We apply the action-availability condition of~\cite{rtc2025} to sequential CPU queries and distinguish lagged from time-aligned execution.

\textbf{Language conditioning.}
Feature-wise linear modulation (FiLM)~\cite{film2018} uses language to scale and shift visual features in BC-Z~\cite{bcz2021}, RT-1~\cite{rt1_2023}, and BAKU~\cite{baku2024}.
MT-ACT~\cite{roboagent2024} combines FiLM in the backbone with a language token in an ACT policy, and Octo~\cite{octo2024} reads frozen T5 tokens.
Concurrent work, DEM~\cite{dem2026}, also caches instruction embeddings from a separate encoder and evaluates decoupled policies on a GPU.
IMPACT combines these established mechanisms while caching instruction-only features to limit the language-processing cost of each CPU query.

\textbf{CPU inference.}
Our micro-kernel follows GotoBLAS and BLIS~\cite{goto2008anatomy,vanzee2015blis}, with register constraints based on Low et al.~\cite{low2016analytical} and empirical validation on each target.
Production runtimes such as XNNPACK~\cite{xnnpack}, ONNX Runtime~\cite{onnxruntime}, OpenVINO~\cite{openvino}, ExecuTorch~\cite{executorch}, and llama.cpp~\cite{llamacpp} select kernels for each instruction set architecture (ISA).
\texttt{vla.cpp}~\cite{vlacpp2026} extends llama.cpp with cached multimodal prefixes and iterative action heads; its reported evaluation uses discrete and embedded GPUs.
SIMD also accelerates motion planning on CPUs~\cite{vamp2024,capt2024}.
\ours{} focuses on CPU execution across policy families, testing how kernel, threading, and precision choices affect action-supply budgets.

% =====================================================================
\section{Action-Supply Budgets}
\label{sec:budget}
% =====================================================================

We first determine when a policy can supply actions continuously despite inference delay.
A controller running at frequency $\fc$ consumes one action every $\Delta=1/\fc$ seconds.
A policy predicts $N$ actions from the latest observation, of which a window of $H\le N$ actions is available for execution.
For example, our Diffusion Policy predicts 64 actions but exposes a 32-action window; the other settings use their full predicted chunk (Table~\ref{tab:policies}).
We index this window from the observation's control step $t$ and denote query latency by $\Tq$, or $d=\lceil\fc\Tq\rceil$ control steps.
The analysis assumes fixed delay, steady-state execution, and sequential queries starting on control ticks $s\ge d$ steps apart.
The ratio $\feff=H/\Tq$, termed effective control rate in~\cite{vlacpp2026}, is an \emph{action-supply rate}; it does not measure feedback frequency.

We consider two execution modes.
In \emph{lagged} execution, the robot starts each chunk from its first action upon arrival.
To avoid running out of actions, the next chunk must arrive before the current one is exhausted, requiring $s \le H$.
In \emph{time-aligned} execution, actions retain their intended control steps, and stale actions are discarded, as in real-time chunking~\cite{rtc2025} and the LeRobot client for our robot~\cite{smolvla2025,lerobot2026}.
At arrival, $\max(0,H-d)$ actions remain, so covering the next arrival requires $s+d \le H$, equivalent to $d \le H-s$ in~\cite{rtc2025}.

Combining these requirements with $s\ge d$, a feasible query interval exists when
\begin{equation}
\text{(C1)}\quad d\le H, \qquad \text{(C2)}\quad 2d\le H,
\label{eq:budget}
\end{equation}
for lagged (C1) and time-aligned (C2) execution, respectively; both can use $s=d$.
C1 is equivalent to $\feff\ge\fc$; ignoring rounding to control steps gives the approximation $\feff\ge2\fc$ for C2.
Thus, time-aligned execution needs enough actions to cover both the current inference delay and the wait for the next chunk.
These conditions ensure action availability under the stated model; smooth chunk transitions and task success require separate evaluation.
Each executed action uses an observation at least $d$ steps old.

We evaluate these nominal budgets using median engine latency, excluding camera capture and transport.
Reliable deployment must also account for these additional delays and occasional slow queries.
We use $\fc=30$\,Hz, matching our SO-101 controller~\cite{so101} and the frame rate of its training datasets.

% =====================================================================
\section{The \ours{} Engine}
\label{sec:engine}
% =====================================================================

\begin{figure*}[t]
\centering
\includegraphics[width=\linewidth]{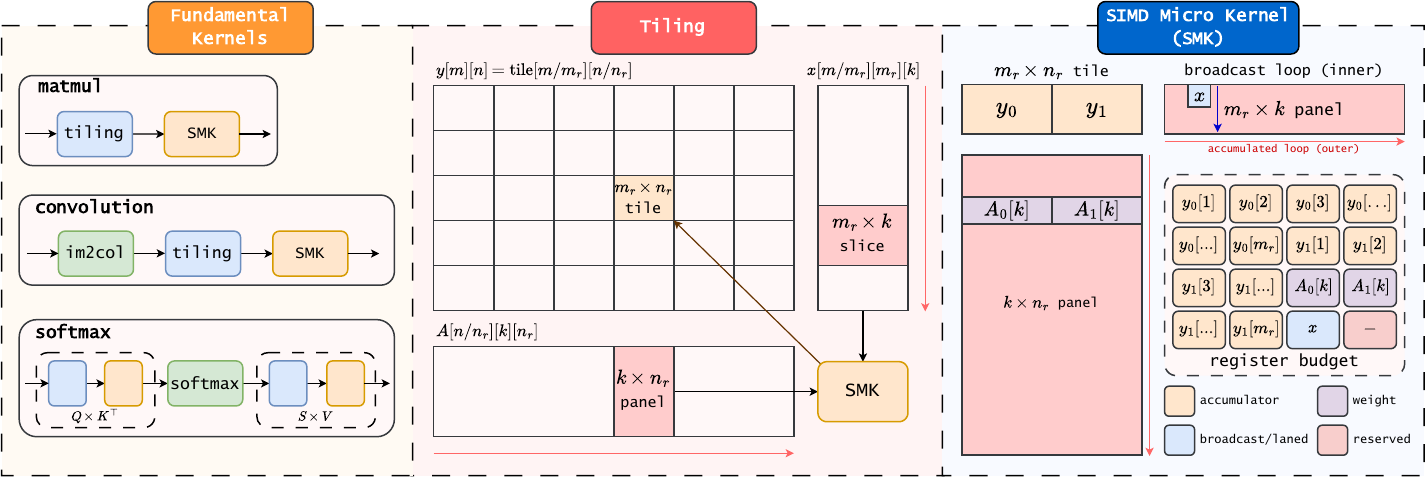}
\caption{SIMD execution in \ours{}.
Linear layers, convolutions, and attention projections share a packed-panel SIMD micro-kernel (SMK).
Convolution supplies image patches to this kernel one panel at a time.
The labels $x$, $y$, and $A$ denote $X$, $Y$, and packed $W^{\top}$, respectively.
The $\mr \times \nr$ output tile remains in registers during accumulation (right).
The attention block uses a separate kernel for $QK^{\top}$ and $SV$, where $S$ is the row-wise softmax of the scaled scores; the key projection writes $K^{\top}$ directly.}
\label{fig:engine}
\end{figure*}

\subsection{Computation Reuse}

\ours maps raw camera frames, joint state, and an instruction to an action chunk in robot units, without a deep learning framework at run time.
It handles resizing, normalization, and tokenization; query timing excludes camera capture, transport, and episode initialization.

Weights are converted offline to contiguous fp32 arrays and packed once per process; instruction-only tensors are cached per episode.
Each query computes observation-dependent features, including image-dependent language features.
Shared operators dispatch to Intel AVX2, AMD AVX2, ARMv8 NEON, or Apple silicon backends.
Apple uses Accelerate~\cite{accelerate} for dense matrix products and convolutions by default; the others use the shared micro-kernel (SMK).

\subsection{Selecting a Register Tile for Each Target}
\label{sec:smk}

A linear layer computes $Y = XW^{\top} + \mathbf{1}_m b^{\top}\in\mathbb{R}^{m\times n}$, where $X\in\mathbb{R}^{m\times k}$, $W\in\mathbb{R}^{n\times k}$, $b\in\mathbb{R}^{n}$, and $\mathbf{1}_m$ is the all-ones vector.
Following GotoBLAS and BLIS~\cite{goto2008anatomy,vanzee2015blis}, $W^{\top}$ is packed into panels of $\nr$ output columns.
The micro-kernel accumulates an $\mr\times\nr$ output tile in vector registers over depth $k$, then adds bias on storage.
Fused multiply-add (FMA) instructions follow increasing depth order, preserving each output's accumulation order across tile dimensions.

With $L$ fp32 values per vector register and $\nv = \nr/L$ registers per tile row, a tile requires $\mr\nv$ accumulator registers, $\nv$ weight registers, and $q$ activation registers.
AVX2 broadcasts each input value into one register, reusing that register across rows, so $q=1$ and each depth step issues $\Lambda = \nv{+}\mr$ loads.
NEON instead keeps one vector per row and reads its lanes over $L$ consecutive depth steps, giving $q=\mr$ and an average of $\Lambda = \nv{+}\mr/L$ loads per step for unpacked activations.
For a target with $\Rmax$ vector registers, throughputs of $\Pf$ vector FMAs and $\Pl$ vector loads per cycle, and FMA latency $\tau$ cycles, we require
\begin{align}
\mr\nv + \nv + q &\le \Rmax, \label{eq:reg}\\
\mr\nv &\ge \Pf\,\tau, \label{eq:lat}\\
\mr\nv\,\Pl &\ge \Lambda\,\Pf, \label{eq:fmab}
\end{align}
Equation~\eqref{eq:reg} limits register use, \eqref{eq:lat} provides enough independent accumulators to hide FMA latency~\cite{low2016analytical}, and \eqref{eq:fmab} requires loads to keep pace with arithmetic.
Selection runs offline: \textsc{Spills} compiles a candidate and checks its inner loop for vector reloads from the stack.

\begin{algorithm}[t]
\caption{Offline register-tile selection for one target}
\label{alg:tile}
\begin{algorithmic}[1]
\Require $L, \Rmax, \Pf, \Pl, \tau$, operand mode, output widths $\mathcal{N}$ of the packed layers
\State $g \gets \gcd \mathcal{N}$; best $\gets \bot$ \Comment{$\bot$: unpacked fallback}
\For{each $\nr$ with $L \mid \nr$ and $\nr \mid g$, ascending}
  \State $\nv \gets \nr/L$
  \For{$\mr \gets \Rmax$ \textbf{down to} $1$}
    \State $(q,\Lambda)\gets\begin{cases}(1,\nv+\mr),&\text{AVX2}\\(\mr,\nv+\mr/L),&\text{NEON}\end{cases}$
    \If{\eqref{eq:reg}--\eqref{eq:fmab} hold \textbf{and not} \Call{Spills}{$\mr,\nr$}}
      \If{best $=\bot$ \textbf{or} $\phi(\mr,\nr) > \phi(\text{best})$}
        \State best $\gets (\mr,\nr)$
      \EndIf
      \State \textbf{break} \Comment{largest feasible $\mr$ for this $\nr$}
    \EndIf
  \EndFor
\EndFor
\State \Return best
\end{algorithmic}
\end{algorithm}

Algorithm~\ref{alg:tile} considers widths $\nr$ divisible by $L$ and dividing the packed layers' greatest common output-width divisor $g$.
The score $\phi(\mr,\nr)=\mr\nr/(\nr+L\mr)$ increases with $\mr$ at fixed $\nr$, justifying the early break; the highest-scoring feasible tile wins, with unpacked weights as fallback.
The score is the AVX2 broadcast kernel's FMA-to-load ratio; its use for NEON is an empirical heuristic.
The NEON load model alone favors $9{\times}8$ and $8{\times}8$, yet the selected $4{\times}16$ tile is 12\% and 19\% faster on the Cortex-A76.

For our policy widths ($g=64$), this procedure selects $6{\times}16$ on AVX2, using 15 of 16 registers, and $4{\times}16$ on NEON, using 24 of 32.
The AVX2 choice matches the single-precision tile used by BLIS for Haswell and Zen.
On the Pi~5, the selected tile reaches 106.9\,GFLOP/s with four threads and 96\% of peak arithmetic throughput on one core with data in L2 cache.
On both ISAs, the selected tile is the fastest tested candidate across all six benchmarked layer shapes.

\subsection{Convolution and Attention}

Convolution reuses the micro-kernel through im2col~\cite{chellapilla2006}: rows contain image patches and packed weights contain flattened filters.
Constructing patches one panel at a time respects stride and padding without storing the full patch matrix.

Attention's query ($Q$), key ($K$), value ($V$), and output projections use the same linear operator; the key projection writes $K^{\top}$ directly by transposing tiles on storage.
The attention products $QK^{\top}$ and $\mathrm{softmax}(QK^{\top}/\sqrt{d_{\mathrm{h}}})V$ use a separate per-backend kernel; $d_{\mathrm{h}}$ is the head width, and softmax operates row-wise over keys.
Both operands change with each query, and the key count need not be a multiple of $\nr$.
Linear layers whose output width is not a multiple of $\nr$, such as action heads, use a row-blocked kernel.

\subsection{Runtime Settings and Quantization}

The backend and register tile are fixed at build time.
Runtime settings control weight packing, convolution panels, matrix-multiplication blocking, attention kernels, and threading (Table~\ref{tab:transfer}).
Optional quantization represents both weights and activations as signed 8-bit integers (W8A8).
This path uses symmetric quantization, 32-bit integer accumulation, and fp32 rescaling, with Arm \texttt{dotprod} or x86 AVX-VNNI instructions for integer products.
Weight scales are per output channel; activation scales are computed dynamically per token for linear layers and per input tensor for convolutions, using the maximum absolute value divided by 127 (scale 1 if zero).
Selected operator groups use this path; normalization, nonlinearities, and unsupported output widths remain fp32.
IMPACT's cached text encoder remains fp32.
Section~\ref{sec:transfer} evaluates these choices per target.

% =====================================================================
\section{IMPACT}
\label{sec:impact}
% =====================================================================

\begin{figure}[t]
\centering
\input{figs/impact-inference.tex}
\caption{IMPACT.
A frozen T5-small encodes the instruction once per episode, providing transformer tokens and FiLM coefficients for each ResNet-18 stage.
Each query runs the modulated backbone and encoder-decoder to predict a 50-step action chunk.}
\label{fig:impact}
\end{figure}
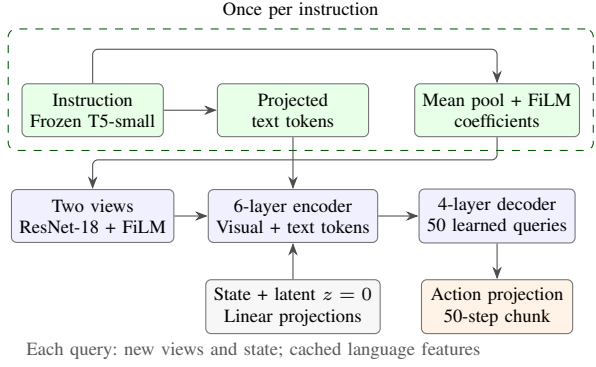

IMPACT (Instruction-Modulated Perception and Action Chunking with Transformers) adds language-based task selection to an action-chunking policy while keeping most language computation outside the repeated inference loop.
Following ACT~\cite{act2023}, it maps two camera views, robot state, and an instruction to a 50-action chunk (Fig.~\ref{fig:impact}).

\paragraph{Visual and state inputs}
The views share an ImageNet-initialized ResNet-18~\cite{resnet2016}; batch normalization is folded into convolutions for inference.
The SO-101 configuration uses $480{\times}640$ images, producing 600 visual tokens across both views, and predicts six-dimensional joint and gripper commands.
The LIBERO configuration uses $256{\times}256$ images, producing 128 visual tokens, with an eight-dimensional end-effector state and seven-dimensional relative actions.
Linear projections map visual features and robot state to width 512.

\paragraph{Language conditioning}
A frozen T5-small text encoder~\cite{t5} maps the instruction to $E\in\mathbb{R}^{n_\ell\times512}$, where $1\le n_\ell\le32$ counts non-padding tokens.
Longer inputs are truncated, and padding is masked within T5.
These states condition the policy in two ways.
First, projected text tokens are appended to the transformer encoder sequence, with padding excluded.
Second, the mean state, $\bar e=n_\ell^{-1}\sum_{j=1}^{n_\ell}E_{j,:}^{\top}$, modulates the feature map $F_i$ after the residual addition and ReLU of each backbone stage $i$:
\begin{equation}
[\gamma_i; \beta_i] = G_i \bar e + b_i, \qquad F_i \leftarrow (1+\gamma_i) \odot F_i + \beta_i ,
\label{eq:film}
\end{equation}
For $c_i$ channels, $G_i\in\mathbb{R}^{2c_i\times512}$ and $b_i\in\mathbb{R}^{2c_i}$ are learned; $[\gamma_i;\beta_i]$ concatenates the channel scales and shifts, broadcast over pixels.
The symbol $\odot$ denotes element-wise multiplication.
Initializing $G_i$ and $b_i$ to zero makes FiLM an identity operation at the start of training.
The engine caches the projected text tokens and FiLM coefficients per instruction and applies the modulation to each query's features.

\paragraph{Action prediction and training}
A six-layer transformer encoder and four-layer decoder, with eight attention heads and feed-forward width 3200, predict the chunk from 50 learned action queries in one pass.
The encoder receives the visual and text tokens, one state token, and one latent token.
Visual tokens use two-dimensional sinusoidal positions; state, latent, and text positions are learned.
Training follows ACT's conditional variational autoencoder (CVAE) formulation: a four-layer encoder receives the demonstrated actions, state, and instruction and predicts a distribution over a 32-dimensional latent $z$.
The objective combines L1 action reconstruction with a Kullback--Leibler penalty of weight 10 toward a standard normal latent prior; inference sets $z=0$.
State and actions are normalized by the training-set mean and standard deviation; predictions are converted back to robot units before execution.
IMPACT has 60\,M parameters in its per-query policy and 35.3\,M in the frozen T5-small encoder, whose outputs are cached per episode.

\paragraph{Computational cost}
For the SO-101 configuration, counting a multiply-add as two floating-point operations gives approximately 88\,GFLOP per query.
Text encoding, projection, and FiLM coefficient generation add about 1.2\,GFLOP once per episode.
Relative to the same six-layer encoder and four-layer decoder without language, up to 32 text tokens add about 2.3\,GFLOP per query and FiLM adds 9\,MFLOP, approximately 2.6\% of the total operation count.
Compared with the shallower ACT in Table~\ref{tab:policies}, IMPACT takes $1.20$--$1.33\times$ as long per query across the four CPUs.

% =====================================================================
\section{Experiments}
\label{sec:exp}
% =====================================================================

\subsection{Setup}

\begin{table}[t]
\centering
\caption{CPU platforms and benchmark settings.
P/E: performance/efficiency cores; C/T: cores/hardware threads.}
\label{tab:devices}
\footnotesize
\setlength{\tabcolsep}{3pt}
\begin{tabular}{lllllr}
\toprule
Device & CPU core & ISA & Cores & Kernel & Thr. \\
\midrule
Apple M4 & Apple & NEON & 4P+6E & Accelerate & 8 \\
Intel i9-14900HX & Raptor Lake & AVX2 & 8P+16E & SMK $6{\times}16$ & 16 \\
AMD Ryzen 5 5500 & Zen 3 & AVX2 & 6C/12T & SMK $6{\times}16$ & 12 \\
Raspberry Pi~5 & Cortex-A76 & NEON & 4C & SMK $4{\times}16$ & 4 \\
\bottomrule
\end{tabular}
\end{table}

\begin{table}[t]
\centering
\caption{Evaluated policies and available action windows $H$.
``+T5'' denotes a separate frozen text encoder.}
\label{tab:policies}
\footnotesize
\setlength{\tabcolsep}{1.5pt}
\begin{tabular}{llllr}
\toprule
Policy & Params & Vision / language & Action model & $H$ \\
\midrule
ACT~\cite{act2023} & 34\,M & ResNet-18 / - & CVAE transformer & 100 \\
DP~\cite{diffusionpolicy2023} & 278\,M & ResNet-18 / - & U-Net, 100/10 steps & 32 \\
Octo-Small~\cite{octo2024} & 27\,M+T5 & conv stem / T5-base & diffusion, 20 steps & 4 \\
TurboVLA~\cite{turbovla2026} & 0.2\,B & DINOv3 / BERT & ACT-style decoder & 12 \\
SmolVLA~\cite{smolvla2025} & 450\,M & SmolVLM2 & flow matching & 50 \\
IMPACT (ours) & 60\,M+T5 & ResNet-18 / T5-small & CVAE transformer & 50 \\
\bottomrule
\end{tabular}
\end{table}

\paragraph{Hardware and policies}
We evaluate six policies on four CPUs, including a passively cooled Pi~5 (Tables~\ref{tab:devices} and~\ref{tab:policies}).
CPU benchmarks use checkpoints fine-tuned on 30\,Hz SO-101 data, except for TurboVLA, whose LIBERO checkpoint is evaluated against the same action-supply budget.
Diffusion Policy uses one checkpoint with 100-step DDPM~\cite{ddpm2020} or 10-step DDIM~\cite{ddim2021} sampling.

\paragraph{Measurement protocol}
We time each query from observation preprocessing through conversion of predicted actions to robot units, as defined in Section~\ref{sec:engine}.
Nominal latency is the median of 15 queries after warmup, measured in a fresh process with the CPU initially at its idle temperature.
To compare settings A and B, we interleave measurements in A--B--B--A order within one session to limit thermal drift.
Before timing fp32 implementations, we check their outputs against the corresponding reference policy; quantization is evaluated separately.

\subsection{Action Supply and Thermal Effects}
\label{sec:rq1}

\begin{table}[t]
\centering
\caption{Action-supply rate $\feff$ (actions/s).
\fail{Gray}: fails C1; \cone{italic}: meets C1 only; upright black: meets C2, at $\fc=30$\,Hz.
Nominal: fp32; last two columns: after a 90\,s Pi~5 soak; dashes: unmeasured.}
\label{tab:rate}
\footnotesize
\setlength{\tabcolsep}{2.6pt}
\begin{tabular}{lrrrrrrrr}
\toprule
& & & & & \multicolumn{4}{c}{Raspberry Pi~5} \\
\cmidrule(l){6-9}
& & & & & \multicolumn{2}{c}{Nominal} & \multicolumn{2}{c}{90\,s soak} \\
\cmidrule(lr){6-7}\cmidrule(l){8-9}
Policy & $H$ & M4 & i9 & Ryzen & $\feff$ & $\Tq$ (s) & fp32 & int8 \\
\midrule
ACT & 100 & 1150 & 892 & 627 & 109 & 0.92 & 89.9 & 242 \\
DP, 100 steps & 32 & \fail{7.2} & \fail{7.2} & \fail{3.6} & \fail{0.8} & 40.1 & -  & -  \\
DP, 10 steps & 32 & \cone{58} & \cone{55} & \fail{29.3} & \fail{6.3} & 5.08 & -  & -  \\
Octo-Small & 4 & 76 & \cone{48} & \cone{48} & \fail{7.8} & 0.51 & \fail{5.8} & \fail{11.3} \\
TurboVLA & 12 & 87 & \cone{58} & \cone{57} & \fail{9.3} & 1.28 & \fail{7.2} & -  \\
SmolVLA & 50 & 73 & \cone{42} & \cone{38} & \fail{6.1} & 8.19 & -  & -  \\
IMPACT & 50 & 453 & 334 & 262 & \cone{41} & 1.21 & \cone{33.5} & 81.2 \\
\bottomrule
\end{tabular}
\end{table}

We test whether each policy supplies enough actions for lagged execution (C1) or for execution that discards stale actions (C2).
To assess the effect of heating, we also measure latency after 90\,s of continuous inference on the passively cooled Pi~5.

Table~\ref{tab:rate} and Fig.~\ref{fig:budget} show that most evaluated policies meet C1 on desktop CPUs, while only ACT and IMPACT qualify on the Pi~5.
IMPACT is the only evaluated language-conditioned policy to supply at least 30 actions/s on the Pi~5.
Octo-Small queries are shorter, but their four-action window is insufficient for either budget.
After this workload, IMPACT supplies 33.5 actions/s in fp32 and 81.2 with int8, satisfying C1 and C2, respectively.

\subsection{Speed and Numerical Fidelity}

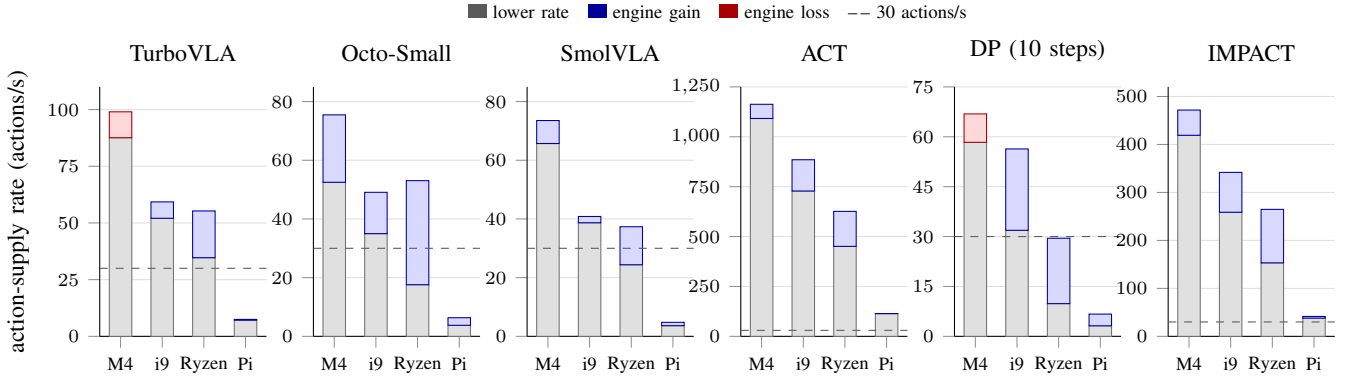
\begin{figure*}[t]
\centering
\input{figs/rate-comparison.tex}
\caption{Paired fp32 action-supply rates for six configurations on four CPUs; panel scales differ.
Gray bars show the lower of the two rates; blue/red extensions show an engine gain/loss relative to \texttt{torch.compile}.
The dashed line marks 30 actions/s; it does not denote feedback frequency.
On the Ryzen, Octo-Small and SmolVLA cross this line with \ours{}.
Diffusion Policy with 100 steps is omitted from the figure.}
\label{fig:rate-speedup}
\end{figure*}

We compare \ours{} with compiled PyTorch using default-mode \texttt{torch.compile}~\cite{pytorch2024} and the Inductor CPU backend.
We also measure uncompiled (eager) execution, using PyTorch 2.11 for five policies and 2.13 for TurboVLA.
All implementations use fp32 and the same query-timing boundary over six interleaved blocks after warmup; compilation time is excluded.
Both implementations cache instruction-only encodings where supported and use the same thread count, with one PyTorch inter-op thread.
We check numerical fidelity against each policy's PyTorch reference using identical observations and stochastic inputs, with preset tolerances of $10^{-2}$ absolute error and $10^{-4}$ relative to action magnitude.
Paired timings are collected in the same session and may differ from the nominal measurements in Table~\ref{tab:rate}.

For each configuration--CPU pair, speedup is the median within-block ratio $T_{\mathrm{compile}}/T_{\mathrm{engine}}$.
The median across 28 pairs (six policies, with two DP samplers, on four CPUs) is approximately $1.4\times$, spanning $0.83$--$3.37\times$.
Eager and compiled execution have similar latency in most settings; Octo-Small on the M4 shows the largest gain from compilation.
On the Ryzen, Octo-Small and SmolVLA cross C1 (Fig.~\ref{fig:rate-speedup}).
The M4 regressions do not change C1 feasibility, although 10-step Diffusion Policy loses its C2 margin.

A separate numerical comparison on 2--12 observations per reported policy gives maximum absolute action differences of at most $8.8{\times}10^{-5}$ in robot units.
These small discrepancies are consistent with changes in floating-point accumulation order.
The Intel and AMD backends produce bitwise-identical fp32 actions on the tested inputs.
For IMPACT, the reference is our PyTorch implementation.
Task performance is evaluated separately below.

\subsection{Processor-Specific Implementation Choices}
\label{sec:transfer}

\begin{table}[t]
\centering
\caption{Effects of individual runtime settings on ACT.
A/B: consistently faster setting; =: no consistent difference.
Values are median $100(T_B/T_A-1)$ (\%); positive favors A.
Bottom: $T_{\mathrm{fp32}}/T_{\mathrm{W8A8}}$; above 1 is faster; n/a: unsupported; --: unmeasured.}
\label{tab:transfer}
\footnotesize
\setlength{\tabcolsep}{2.1pt}
\begin{tabular}{llrrrr}
\toprule
Setting & A\,/\,B & M4 & i9 & Ryzen & Pi~5 \\
\midrule
Attention kernel & library / custom & A\,$+$9.9 & = & = & = \\
bf16 expansion & off / on & = & = & = & = \\
Min. panel rows & 16 / 4 & = & = & B\,$-$4.2 & = \\
Convolution & tiled / flat & = & A\,$+$156 & A\,$+$260 & A\,$+$16.9 \\
Block rows & auto / 64 & = & A\,$+$8.3 & A\,$+$5.0 & = \\
Thread schedule & static / dyn. & = & = & = & = \\
Parallel cutoff & 8192 / 1024 & = & = & = & = \\
Weight packing & on / off & = & A\,$+$31.0 & A\,$+$30.7 & A\,$+$14.8 \\
Thread count & more / fewer & = & A\,$+$5.4 & = & A\,$+$32.0 \\
Threads per view & auto / 2 & B\,$-$6.2 & A\,$+$104 & A\,$+$75.6 & A\,$+$17.8 \\
Thread waiting & passive / active & A\,$+$72.9 & A\,$+$1.7 & = & = \\
\midrule
\multicolumn{2}{l}{W8A8, ACT} & 1.13 & 1.24 & n/a & 2.68 \\
\multicolumn{2}{l}{W8A8, IMPACT} & 1.06 & 1.26 & n/a & 2.39 \\
\multicolumn{2}{l}{W8A8, Octo-Small} & 0.66 & 1.03 & n/a & 1.73 \\
\multicolumn{2}{l}{W8A8, SmolVLA} & 0.71 & 1.14 & n/a & -  \\
\bottomrule
\end{tabular}
\end{table}

We vary eleven settings individually on ACT using the paired protocol above.
We report a consistent improvement only when a setting is faster in all but at most one pair and its median effect exceeds baseline drift.
In Table~\ref{tab:transfer}, bf16 expansion converts bfloat16 weights to fp32; the parallel cutoff is the minimum element count for parallel execution.
The higher/lower thread counts are 10/4 on the M4, 32/8 on the i9, 12/6 on the Ryzen, and 4/2 on the Pi~5.
We also compare W8A8 with fp32 on supported policy--CPU pairs and examine selective quantization of convolution and attention layers.
The W8A8 comparisons quantize all supported operator groups: for IMPACT, these are ResNet convolutions, token projections, and encoder/decoder linear layers, with the text encoder kept in fp32.
Varying one setting at a time measures its individual effect; interactions between settings are not evaluated.

Table~\ref{tab:transfer} shows that useful settings depend on the target CPU: tiled convolution benefits the non-Apple backends, while extra camera-view threads help the M4 but hurt the other devices.
Quantization is similarly uneven, yielding substantial gains on the Pi~5 but slowing some policies on the M4.
For IMPACT on the M4, quantizing only convolutions is faster than quantizing all layers.
These results support selecting settings and quantized layers through measurements on each target.

\subsection{Manipulation and Instruction Following}

\begin{table}[t]
\centering
\caption{LIBERO success rates (\%); protocols differ across published baselines.
IMPACT: final checkpoint versus selection on the same evaluation episodes.}
\label{tab:libero}
\footnotesize
\setlength{\tabcolsep}{2.4pt}
\begin{tabular}{lrrrrr}
\toprule
Policy & Spatial & Object & Goal & Long & Avg \\
\midrule
\multicolumn{6}{l}{\emph{No robot pretraining}} \\
Diffusion Policy & 78.3 & 92.5 & 68.3 & 50.5 & 72.4 \\
TurboVLA & 99.2 & 99.8 & 97.4 & 94.2 & 97.7 \\
SmolVLA & 90.0 & 96.0 & 92.0 & 71.0 & 87.3 \\
IMPACT, final & 83.5 & 83.5 & 84.5 & 54.0 & 76.4 \\
IMPACT, selected & 83.5 & 83.5 & 89.0 & 58.5 & 78.6 \\
\midrule
\multicolumn{6}{l}{\emph{Pretrained on robot data}} \\
Octo-Base & 78.9 & 85.7 & 84.6 & 51.1 & 75.1 \\
OpenVLA & 84.7 & 88.4 & 79.2 & 53.7 & 76.5 \\
\bottomrule
\end{tabular}
\end{table}

\paragraph{LIBERO}
We evaluate IMPACT on four LIBERO simulation suites~\cite{libero2023}, each containing ten manipulation tasks.
We train one policy per suite without robot pretraining and evaluate each task from 20 initial states on a GPU host.
Training uses AdamW with batch size 64, seed 1000, constant learning rate $10^{-4}$, and weight decay $10^{-4}$.
Evaluation uses two $256{\times}256$ views and executes all 50 relative end-effector actions before querying the policy again.
The final 30,000-step checkpoints achieve 76.4\% mean success (Table~\ref{tab:libero}).
Selecting checkpoints on these same evaluation rollouts yields 78.6\%; we report this exploratory result separately because selection can inflate the estimate.
Published DP, Octo-Base, and OpenVLA results average three seeds of 500 rollouts per suite; TurboVLA uses 500 and SmolVLA 100~\cite{openvlaoft2025,turbovla2026,smolvla2025}.
These published results use different observation inputs and execution horizons and provide context rather than a controlled comparison.

\paragraph{Instruction following}
To test whether IMPACT uses language to select a task, we keep the LIBERO-Goal environments fixed and change only the instruction, using the selected 25,000-step checkpoint.
All ten tasks share the same scene and objects.
We evaluate three instruction permutations, each replacing every original instruction, for 600 rollouts in total.
We score both the original and newly instructed goals using LIBERO's success conditions.
IMPACT achieves the newly instructed goal in 91\% of trials and the original goal in none, against an 88.5\% correct-instruction control.
This intervention supports language-based selection among familiar goals; generalization to unseen wording and the individual contributions of the two conditioning paths remain untested.

\begin{figure}[t]
\centering
\includegraphics[width=\linewidth]{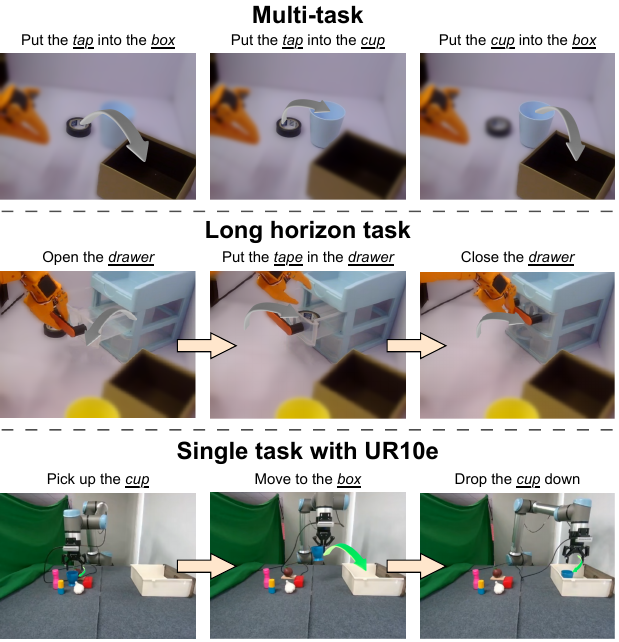}
\caption{Real-robot tasks on two embodiments.
Top: three instructions in a shared SO-101~\cite{so101} scene vary the destination or grasp target.
Middle: the SO-101 opens a drawer, places the tape inside, and closes it.
Bottom: a UR10e with a Robotiq gripper picks up the cup and places it in the box.
The middle and bottom rows show successive stages from left to right.}
\label{fig:so101}
\end{figure}

\paragraph{SO-101}
On the SO-101 arm (Fig.~\ref{fig:so101}), we use 44 demonstrations for three instructions in a shared scene and 20 for a separate drawer task.
LeRobot's asynchronous interface~\cite{lerobot2026} connects to \ours{} on the M4, Pi~5, and i5-12400F, or to PyTorch on a GPU.
The trials use one scene and one operator; IMPACT results pool trials across these CPU and GPU servers.
Each row of Table~\ref{tab:robot} reports 20 trials with a Wilson 95\% confidence interval.
IMPACT performs better on destination selection than on changing the grasp target (Table~\ref{tab:robot}).
The fp32 and int8 success counts are similar, although 20 trials per condition are insufficient to establish equivalent performance.

As a no-language control, we train ACT with IMPACT's chunk length, transformer depth, and shared ResNet-18, removing text and FiLM.
This control is deeper than the ACT in the CPU benchmarks.
We train one policy on all three tasks and three specialist policies on one task each.
All train for 19,000 steps with batch size 8 and execute synchronously, so inference and motion do not overlap.
Each specialist therefore receives more passes over its smaller dataset.
The multi-task control succeeds in 4 of 20 trials, compared with 46 of 60 across specialists.
The operator attributes 9 of the 20 multi-task trials to target ambiguity and 7 to grasp failures.
These observations suggest task ambiguity as a failure mode, but differences in data exposure and execution protocol prevent attributing the success gap to language conditioning alone.

\paragraph{UR10e}
SmolVLA, served by \ours{} from the M4 and Ryzen, places a cup into a box using a UR10e with a Robotiq gripper (Table~\ref{tab:ur10e}).
Both hosts achieve 12 successes in 20 trials, with a $2.0\times$ difference in mean query round-trip time.
The small samples leave the effect of host latency unresolved.
Round-trip times include transport and serialization, exceeding nominal engine latency by approximately 0.5\,s on the M4 and 1.0\,s on the Ryzen.
These additional delays show why deployment must be assessed using the full observation-to-action path.

\begin{table}[t]
\centering
\caption{SO-101 success rates (SR, \%): 20 trials/row, Wilson 95\% confidence intervals (CI).
IMPACT pools servers; ACT controls execute synchronously.}
\label{tab:robot}
\footnotesize
\setlength{\tabcolsep}{3pt}
\begin{tabular}{llll}
\toprule
Model & Goal & Training & SR (\%, 95\% CI) \\
\midrule
ACT         & all three                 & multi-task  & 20 (8--42) \\
ACT         & tape $\rightarrow$ box    & single-task & 60 (39--78) \\
ACT         & tape $\rightarrow$ cup    & single-task & 90 (70--97) \\
ACT         & cup $\rightarrow$ box     & single-task & 80 (58--92) \\
\midrule
IMPACT      & tape $\rightarrow$ box    & multi-task  & 90 (70--97) \\
IMPACT      & tape $\rightarrow$ cup    & multi-task  & 85 (64--95) \\
IMPACT      & cup $\rightarrow$ box     & multi-task  & 60 (39--78) \\
\midrule
IMPACT int8 & tape $\rightarrow$ box    & multi-task  & 85 (64--95) \\
IMPACT int8 & tape $\rightarrow$ cup    & multi-task  & 95 (76--99) \\
IMPACT int8 & cup $\rightarrow$ box     & multi-task  & 50 (30--70) \\
\midrule
IMPACT      & tape $\rightarrow$ drawer & drawer      & 65 (43--82) \\
\bottomrule
\end{tabular}
\end{table}

\begin{table}[t]
\centering
\caption{SmolVLA on UR10e: 20 trials/host, Wilson 95\% CI.
Mean query round-trip times include transport.}
\label{tab:ur10e}
\footnotesize
\setlength{\tabcolsep}{3pt}
\begin{tabular}{lrl}
\toprule
Host & Round-trip (ms) & SR (\%, 95\% CI) \\
\midrule
Apple M4     & 1177 & 60 (39--78) \\
Ryzen 5 5500 & 2343 & 60 (39--78) \\
\bottomrule
\end{tabular}
\end{table}

% =====================================================================
\section{Conclusion}
% =====================================================================

We presented \ours{}, an inference engine that combines shared SIMD kernels with target-specific optimization, and IMPACT, an ACT-based policy that adds language-guided task selection with little repeated computation.
CPU deployment becomes feasible when the execution horizon covers inference delay and the implementation uses the processor efficiently.
On the Ryzen, \ours{} pushes Octo-Small and SmolVLA across the 30\,Hz threshold.
On the Pi~5, IMPACT is the only language-conditioned policy in our evaluated set that meets this budget, and its int8 path also meets the stricter time-aligned budget after a thermal soak.
Instruction shuffling and real-robot trials demonstrate language-based goal selection, extending ACT to instruction-guided manipulation.

\section{Limitations}

Our evaluation covers policies up to 450\,M parameters and compares CPU inference with eager and compiled PyTorch; it does not benchmark GPUs, embedded accelerators, or dedicated runtimes such as ONNX Runtime and OpenVINO.
The budgets use median engine latency, whereas uninterrupted deployment depends on the tail of the full observation-to-action delay.
Thermal behavior is measured over one 90\,s soak.
The int8 path is validated on the SO-101 multi-task bench, but its effect on simulation success and on the long-horizon task is unmeasured.
LIBERO evaluation uses one seed, and the SO-101 trials pool servers, limiting conclusions about behavior on individual CPUs.

\bibliographystyle{IEEEtran}
\bibliography{IEEEabrv}

\end{document}

%% file: figs/budget.tex
% Fig. 1: nominal action-supply budgets. Requires pgfplots.
% x: median query latency (ms), y: actions per query H.
% Rows sharing a horizon (H=50, H=32) are drawn at H*1.12 and H/1.12.
\begin{tikzpicture}[font=\footnotesize]
% Okabe-Ito based palette.
\definecolor{cACT}{HTML}{0072B2}
\definecolor{cIMP}{HTML}{D55E00}
\definecolor{cSmol}{HTML}{AA3377}
\definecolor{cDP}{HTML}{009E73}
\definecolor{cTurbo}{HTML}{7B61C9}
\definecolor{cOcto}{HTML}{3A9BD5}
\definecolor{ink}{HTML}{333333}
% filled five-point star
\pgfdeclareplotmark{fstar}{%
  \pgfpathmoveto{\pgfpointpolar{90}{1.45\pgfplotmarksize}}%
  \foreach \a in {1,...,9}{%
    \pgfmathsetmacro\r{mod(\a,2)==0 ? 1.45 : 0.6}%
    \pgfpathlineto{\pgfpointpolar{90+36*\a}{\r\pgfplotmarksize}}}%
  \pgfpathclose\pgfusepathqfillstroke}
\begin{axis}[
  width=\columnwidth, height=5.2cm,
  xmode=log, ymode=log,
  xmin=20, xmax=60000, ymin=2.8, ymax=300,
  xtick={30,100,300,1000,3000,10000,30000},
  ytick={4,12,32,50,100},
  log ticks with fixed point,
  /pgf/number format/1000 sep={\,},
  minor tick num=0, yminorticks=false,
  tick style={black!60}, tick align=outside, tick pos=left,
  axis line style={black!60},
  xlabel={median query latency $\Tq$ (ms)},
  ylabel={available action window $H$},
  xlabel style={yshift=2pt}, ylabel style={yshift=-3pt},
  axis on top,
  scatter/classes={m4={mark=*}, i9={mark=square*}, rz={mark=diamond*,mark size=2.6pt}, pi={mark=triangle*,mark size=2.5pt}},
  every axis plot/.append style={only marks, mark size=2pt, scatter, scatter src=explicit symbolic,
    mark options={draw=white, line width=0.35pt}},
  legend style={at={(0.5,1.03)}, anchor=south, font=\scriptsize, legend columns=6, draw=none,
    inner sep=0pt, /tikz/every even column/.append style={column sep=6pt}},
  legend cell align=left,
]
% regions: meets C1 and C2 (white), C1 only, misses C1
\fill[black!7]  (axis cs:46.667,2.8) -- (axis cs:93.333,2.8) -- (axis cs:10000,300) -- (axis cs:5000,300) -- cycle;
\fill[black!14] (axis cs:93.333,2.8) -- (axis cs:60000,2.8) -- (axis cs:60000,300) -- (axis cs:10000,300) -- cycle;
\draw[black!55, line width=0.6pt] (axis cs:46.667,2.8) -- (axis cs:5000,300)
  node[pos=0.868, sloped, below, inner sep=1.5pt, font=\scriptsize, text=ink] {C2: $H\,{=}\,2\fc\Tq$};
\draw[black!55, line width=0.6pt] (axis cs:93.333,2.8) -- (axis cs:10000,300)
  node[pos=0.868, sloped, below, inner sep=1.5pt, font=\scriptsize, text=ink] {C1: $H\,{=}\,\fc\Tq$};
\node[anchor=north west, font=\scriptsize\itshape, text=black!65] at (rel axis cs:0.01,0.985) {meets C1 \& C2};
\node[anchor=south east, font=\scriptsize\itshape, text=black!65] at (rel axis cs:0.99,0.015) {below C1};
% device legend
\addlegendimage{only marks, mark=*, ink} \addlegendentry{M4}
\addlegendimage{only marks, mark=square*, ink} \addlegendentry{i9}
\addlegendimage{only marks, mark=diamond*, mark size=2.6pt, ink} \addlegendentry{Ryzen}
\addlegendimage{only marks, mark=triangle*, mark size=2.5pt, ink} \addlegendentry{Pi\,5}
\addlegendimage{only marks, mark=fstar, mark size=2.2pt, ink} \addlegendentry{Pi\,5 int8}
% data (M4, i9, Ryzen, Pi 5)
\addplot[forget plot, cACT] coordinates {(86.9,100)[m4] (112.1,100)[i9] (159.5,100)[rz] (915.9,100)[pi]};
\addplot[forget plot, cSmol] coordinates {(685.5,44.643)[m4] (1182.8,44.643)[i9] (1319.9,44.643)[rz] (8187.5,44.643)[pi]};
\addplot[forget plot, cDP, mark options={fill=cDP!40!white, draw=cDP, line width=0.5pt}]
  coordinates {(4445.7,35.84)[i9] (4449.6,35.84)[m4] (9008.0,35.84)[rz] (40054.8,35.84)[pi]};
\addplot[forget plot, cDP] coordinates {(551.7,28.571)[m4] (584.3,28.571)[i9] (1091.2,28.571)[rz] (5077.8,28.571)[pi]};
\addplot[forget plot, cTurbo] coordinates {(137.7,12)[m4] (207.1,12)[i9] (212.4,12)[rz] (1284.9,12)[pi]};
\addplot[forget plot, cOcto] coordinates {(52.9,4)[m4] (84.0,4)[i9] (83.8,4)[rz] (513.7,4)[pi]};
% IMPACT: cool points only (the 90 s soak markers are not shown)
\draw[->, >=stealth, cIMP, line width=0.5pt, shorten >=4pt, shorten <=3.5pt]
  (axis cs:1212.3,56) to[bend right=45] node[midway, above, inner sep=1pt, font=\scriptsize, text=ink] {int8} (axis cs:551,56);
\addplot[forget plot, cIMP, mark size=2.4pt] coordinates {(110.3,56)[m4] (149.6,56)[i9] (190.8,56)[rz] (1212.3,56)[pi]};
\addplot[forget plot, cIMP, scatter=false, mark=fstar, mark size=2.4pt] coordinates {(551,56)};
% policy labels
\begin{scope}[every node/.style={inner sep=1pt, text=ink}]
\node[anchor=east] at (axis cs:78,100) {ACT};
\node[anchor=east] at (axis cs:98,56) {\textbf{IMPACT}};
\node[anchor=west] at (axis cs:9300,44.643) {SmolVLA};
\node[anchor=north east, align=right] at (axis cs:48000,31.5) {DP\\(100 steps)};
\node[anchor=east] at (axis cs:440,28.571) {DP (10 steps)};
\node[anchor=west] at (axis cs:1450,12) {TurboVLA};
\node[anchor=west] at (axis cs:580,4) {Octo-Small};
\end{scope}
\end{axis}
\end{tikzpicture}

%% file: figs/impact-inference.tex
% Inference graph. Training-only CVAE encoder omitted; its latent is zero here.
\begin{tikzpicture}[
 font=\scriptsize, >=Stealth,
 box/.style={draw=black!55, rounded corners=2pt, align=center,
             minimum height=7mm, inner sep=3pt},
 cached/.style={box, fill=green!9},
 live/.style={box, fill=blue!6},
 arrow/.style={->, draw=black!70, rounded corners=2pt},
]
\node[cached, minimum width=17mm] (text) at (0,0) {Instruction\\Frozen T5-small};
\node[cached, minimum width=20mm] (tokens) at (2.65,0) {Projected\\text tokens};
\node[cached, minimum width=20mm] (film) at (5.35,0) {Mean pool + FiLM\\coefficients};
\draw[arrow] (text) -- (tokens);
\draw[arrow] (text.north) -- ++(0,0.45) -| (film.north);

\node[live, minimum width=20mm] (vision) at (0,-1.4) {Two views\\ResNet-18 + FiLM};
\node[live, minimum width=20mm] (encoder) at (2.65,-1.4) {6-layer encoder\\Visual + text tokens};
\node[live, minimum width=20mm] (decoder) at (5.35,-1.4) {4-layer decoder\\50 learned queries};
\draw[arrow] (vision) -- (encoder);
\draw[arrow] (tokens) -- (encoder);
\draw[arrow] (encoder) -- (decoder);
\draw[arrow] (film.south) -- ++(0,-0.3) -| (vision.north);

\node[box, fill=black!3, minimum width=20mm] (state) at (2.65,-2.6) {State + latent $z=0$\\Linear projections};
\node[box, fill=orange!10, minimum width=20mm] (action) at (5.35,-2.6) {Action projection\\50-step chunk};
\draw[arrow] (state) -- (encoder);
\draw[arrow] (decoder) -- (action);

\begin{scope}[on background layer]
\coordinate (cachetop) at (2.65,0.9);
\node[draw=green!35!black, dashed, rounded corners=3pt, inner sep=5pt,
      fit=(text)(tokens)(film)(cachetop), label={[font=\scriptsize]above:Once per instruction}] {};
\end{scope}
\node[anchor=north west, text=black!65] at (-1,-2.95) {Each query: new views and state; cached language features};
\end{tikzpicture}

%% file: figs/rate-comparison.tex
% Rates are H / median deployment latency from generated/e5_latency.tex.
% Six configurations shown; DDPM-100 remains in the aggregate reported in the text.
\begin{tikzpicture}
\begin{groupplot}[
 group style={group size=6 by 1, horizontal sep=0.035\textwidth},
 scale only axis, width=0.124\textwidth, height=3.3cm,
 xmin=0.5, xmax=4.5, ymin=0,
 xtick={1,2,3,4}, xticklabels={M4,i9,Ryzen,Pi},
 tick label style={font=\scriptsize}, title style={font=\small},
 axis lines*=left, tick align=outside, tick style={black!45},
 ymajorgrids=true, grid style={black!12},
 scaled y ticks=false,
]
\nextgroupplot[title={TurboVLA}, ymax=110, ytick={0,25,...,110}, ylabel={action-supply rate (actions/s)}, ylabel style={font=\small}]
\path[fill=black!12,draw=black!55] (axis cs:0.73,0) rectangle (axis cs:1.27,87.5912);
\path[fill=red!15,draw=red!65!black] (axis cs:0.73,87.5912) rectangle (axis cs:1.27,99.0099);
\path[fill=black!12,draw=black!55] (axis cs:1.73,0) rectangle (axis cs:2.27,52.0607);
\path[fill=blue!15,draw=blue!65!black] (axis cs:1.73,52.0607) rectangle (axis cs:2.27,59.2885);
\path[fill=black!12,draw=black!55] (axis cs:2.73,0) rectangle (axis cs:3.27,34.6821);
\path[fill=blue!15,draw=blue!65!black] (axis cs:2.73,34.6821) rectangle (axis cs:3.27,55.2995);
\path[fill=black!12,draw=black!55] (axis cs:3.73,0) rectangle (axis cs:4.27,7.1344);
\path[fill=blue!15,draw=blue!65!black] (axis cs:3.73,7.1344) rectangle (axis cs:4.27,7.4752);
\draw[dashed,black!65] (axis cs:0.5,30) -- (axis cs:4.5,30);
\nextgroupplot[title={Octo-Small}, ymax=85, ytick={0,20,...,85}]
\path[fill=black!12,draw=black!55] (axis cs:0.73,0) rectangle (axis cs:1.27,52.4934);
\path[fill=blue!15,draw=blue!65!black] (axis cs:0.73,52.4934) rectangle (axis cs:1.27,75.4717);
\path[fill=black!12,draw=black!55] (axis cs:1.73,0) rectangle (axis cs:2.27,34.9956);
\path[fill=blue!15,draw=blue!65!black] (axis cs:1.73,34.9956) rectangle (axis cs:2.27,49.0798);
\path[fill=black!12,draw=black!55] (axis cs:2.73,0) rectangle (axis cs:3.27,17.5901);
\path[fill=blue!15,draw=blue!65!black] (axis cs:2.73,17.5901) rectangle (axis cs:3.27,53.0504);
\path[fill=black!12,draw=black!55] (axis cs:3.73,0) rectangle (axis cs:4.27,3.7636);
\path[fill=blue!15,draw=blue!65!black] (axis cs:3.73,3.7636) rectangle (axis cs:4.27,6.3462);
\draw[dashed,black!65] (axis cs:0.5,30) -- (axis cs:4.5,30);
\nextgroupplot[title={SmolVLA}, ymax=85, ytick={0,20,...,85}]
\path[fill=black!12,draw=black!55] (axis cs:0.73,0) rectangle (axis cs:1.27,65.7030);
\path[fill=blue!15,draw=blue!65!black] (axis cs:0.73,65.7030) rectangle (axis cs:1.27,73.5402);
\path[fill=black!12,draw=black!55] (axis cs:1.73,0) rectangle (axis cs:2.27,38.7027);
\path[fill=blue!15,draw=blue!65!black] (axis cs:1.73,38.7027) rectangle (axis cs:2.27,40.8430);
\path[fill=black!12,draw=black!55] (axis cs:2.73,0) rectangle (axis cs:3.27,24.3926);
\path[fill=blue!15,draw=blue!65!black] (axis cs:2.73,24.3926) rectangle (axis cs:3.27,37.3469);
\path[fill=black!12,draw=black!55] (axis cs:3.73,0) rectangle (axis cs:4.27,3.6335);
\path[fill=blue!15,draw=blue!65!black] (axis cs:3.73,3.6335) rectangle (axis cs:4.27,4.7769);
\draw[dashed,black!65] (axis cs:0.5,30) -- (axis cs:4.5,30);
\nextgroupplot[title={ACT}, ymax=1250, ytick={0,250,...,1250}]
\path[fill=black!12,draw=black!55] (axis cs:0.73,0) rectangle (axis cs:1.27,1091.7031);
\path[fill=blue!15,draw=blue!65!black] (axis cs:0.73,1091.7031) rectangle (axis cs:1.27,1162.7907);
\path[fill=black!12,draw=black!55] (axis cs:1.73,0) rectangle (axis cs:2.27,728.3321);
\path[fill=blue!15,draw=blue!65!black] (axis cs:1.73,728.3321) rectangle (axis cs:2.27,884.9558);
\path[fill=black!12,draw=black!55] (axis cs:2.73,0) rectangle (axis cs:3.27,450.8566);
\path[fill=blue!15,draw=blue!65!black] (axis cs:2.73,450.8566) rectangle (axis cs:3.27,626.1741);
\path[fill=black!12,draw=black!55] (axis cs:3.73,0) rectangle (axis cs:4.27,114.0641);
\path[fill=blue!15,draw=blue!65!black] (axis cs:3.73,114.0641) rectangle (axis cs:4.27,114.2074);
\draw[dashed,black!65] (axis cs:0.5,30) -- (axis cs:4.5,30);
\nextgroupplot[title={DP (10 steps)}, ymax=75, ytick={0,15,...,75}]
\path[fill=black!12,draw=black!55] (axis cs:0.73,0) rectangle (axis cs:1.27,58.3303);
\path[fill=red!15,draw=red!65!black] (axis cs:0.73,58.3303) rectangle (axis cs:1.27,66.8757);
\path[fill=black!12,draw=black!55] (axis cs:1.73,0) rectangle (axis cs:2.27,31.8757);
\path[fill=blue!15,draw=blue!65!black] (axis cs:1.73,31.8757) rectangle (axis cs:2.27,56.3479);
\path[fill=black!12,draw=black!55] (axis cs:2.73,0) rectangle (axis cs:3.27,9.8347);
\path[fill=blue!15,draw=blue!65!black] (axis cs:2.73,9.8347) rectangle (axis cs:3.27,29.4931);
\path[fill=black!12,draw=black!55] (axis cs:3.73,0) rectangle (axis cs:4.27,3.1695);
\path[fill=blue!15,draw=blue!65!black] (axis cs:3.73,3.1695) rectangle (axis cs:4.27,6.6863);
\draw[dashed,black!65] (axis cs:0.5,30) -- (axis cs:4.5,30);
\nextgroupplot[title={IMPACT}, ymax=520, ytick={0,100,...,520}]
\path[fill=black!12,draw=black!55] (axis cs:0.73,0) rectangle (axis cs:1.27,419.1115);
\path[fill=blue!15,draw=blue!65!black] (axis cs:0.73,419.1115) rectangle (axis cs:1.27,471.6981);
\path[fill=black!12,draw=black!55] (axis cs:1.73,0) rectangle (axis cs:2.27,258.6653);
\path[fill=blue!15,draw=blue!65!black] (axis cs:1.73,258.6653) rectangle (axis cs:2.27,341.7635);
\path[fill=black!12,draw=black!55] (axis cs:2.73,0) rectangle (axis cs:3.27,153.1863);
\path[fill=blue!15,draw=blue!65!black] (axis cs:2.73,153.1863) rectangle (axis cs:3.27,264.6903);
\path[fill=black!12,draw=black!55] (axis cs:3.73,0) rectangle (axis cs:4.27,38.0633);
\path[fill=blue!15,draw=blue!65!black] (axis cs:3.73,38.0633) rectangle (axis cs:4.27,41.5076);
\draw[dashed,black!65] (axis cs:0.5,30) -- (axis cs:4.5,30);
\end{groupplot}
\node[anchor=south,font=\scriptsize] at ([yshift=21pt]$(group c1r1.north west)!0.5!(group c6r1.north east)$)
 {\textcolor{black!65}{\rule{7pt}{5pt}} lower rate\quad
  \textcolor{blue!60!black}{\rule{7pt}{5pt}} engine gain\quad
  \textcolor{red!65!black}{\rule{7pt}{5pt}} engine loss\quad
  --\,-- 30 actions/s};
\end{tikzpicture}